\documentclass[10pt,twocolumn,letterpaper]{article}

\usepackage{iccv}
\usepackage{times}
\usepackage{epsfig}
\usepackage{graphicx}
\usepackage{amsmath}
\usepackage{amssymb}

\usepackage[pagebackref=true,breaklinks=true,letterpaper=true,colorlinks,bookmarks=false]{hyperref}

\iccvfinalcopy 

\def\iccvPaperID{19} 

\ificcvfinal\fi
\begin{document}

\title{Improved Descriptors for Patch Matching and Reconstruction}

\author{Rahul Mitra\\
Indian Institute of Technology Bombay\\
Mumbai, India\\
{\tt\small rmitter@cse.iitb.ac.in}
\and
Jiakai Zhang\\
New York University\\
NYC, USA\\
{\tt\small zhjk@nyu.edu}
\and
Sanath Narayan, Shuaib Ahmed\\
Mercedes-Benz Research and Development India Pvt. Ltd.\\
Bangalore, India\\
{\tt\small \{sanath.narayan,shuaib.ahmed\}@daimler.com}
\and
Sharat Chandran, Arjun Jain\\
Indian Institute of Technology Bombay\\
Mumbai, India\\
{\tt\small \{sharat,ajain\}@cse.iitb.ac.in}
}

\maketitle

\begin{abstract}
We propose a convolutional neural network (ConvNet) based approach for learning local image descriptors which can be used for significantly  improved patch matching and 3D reconstructions.  A multi-resolution  
ConvNet is used for learning keypoint descriptors. We also propose a new dataset consisting of an order of magnitude more number of scenes, 
images, and positive and negative correspondences compared to the currently available Multi-View Stereo (MVS)~\cite{mvs-new} dataset. 
The new dataset also has better coverage of the overall viewpoint, scale, and lighting changes in comparison to 
the MVS dataset. We evaluate our approach on publicly available datasets, such as Oxford Affine Covariant Regions Dataset (ACRD) ~\cite{oxford}, MVS~\cite{mvs-new}, Synthetic~\cite{generated-dataset} and \emph{Strecha}~\cite{Strecha} datasets to quantify the image descriptor performance. Scenes from the Oxford~ACRD, MVS and Synthetic datasets are used for evaluating the patch matching performance of the learnt descriptors while the \emph{Strecha} dataset is used to evaluate the 3D reconstruction task. Experiments show that the proposed descriptor outperforms the current state-of-the-art descriptors in both the evaluation tasks.
\end{abstract}

\section{Introduction}{\label{sec-intro}}
Designing high quality descriptors for finding correspondences between images is crucial for many computer vision 
tasks such as 3D reconstruction, structure from motion (SFM) ~\cite{photo-tourism}, wide-baseline matching~\cite{Strecha},
stitching image panoramas~\cite{panorama}, and tracking~\cite{he2009surf, surf}. Finding correspondences in-the-wild is challenging due to changes in viewpoints, scale variations, variations in illumination, occlusion, and shading.

Traditional handcrafted descriptors ~\cite{sift, surf} encode pixel, super-pixel or sub-pixel level statistics and similarity, but do not have ability to capture higher structural level information. However, there are tasks which are highly  dependent on pixel level statistics. In these kind of tasks handcrafted features perform better. Resurgence of ConvNets has resulted in many recent works proposing learning based descriptors~\cite{deepdesc, deepcompare, matchnet, tfeat}. ConvNet based descriptors have the potential to capture higher level structural information and generalize well, if it is properly trained with a good dataset.

As noted in~\cite{tfeat}, current benchmark datasets limit the potential of ConvNet based learning algorithm to evaluate across different datasets. The frequently used datasets for patch matching are the Multi-View Stereo (MVS) dataset~\cite{mvs-new} and Oxford~ACRD dataset~\cite{oxford}. The MVS dataset has only three scenes (each scene consists of approximately $250$ images) and does not provide sufficient variation in terms of scene content, viewpoint, and scale. Further, most of the non-matching pairs in the dataset are totally distinct from each other which seldom happens in real-world scenarios. The Oxford~ACRD dataset which was created a decade ago is very small for today's computing power and is prone to over-fitting and in turn cannot generalize any descriptor to be robust in-the-wild. Even a recently published dataset named \emph{Hpatches}~\cite{hpatches} contains scenes with variations only in illumination and viewpoints on flat surfaces such as walls. Such type of scenes do not suffer from occlusions. However, scenes capturing real world 3D non-planer objects at various angles will experience partial occlusions. Hence, a good dataset should include these characteristics to be more challenging and to efficently describe feature descriptors for 3D reconstruction of non-planer objects.

For efficient ConvNet based descriptors, it is important to have a good combination of ConvNet architecture and dataset on which the ConvNet is trained. Selection of a good architecture that is robust to geometric and scale variations is as essential as good datasets. Working on these lines, in this paper, we propose a multi-resolution ConvNet architecture based descriptors. The ConvNet is trained on a new larger dataset which has higher geometric and photometric variations in the scene, number of viewpoints, variations in scale, and also includes scenes capturing 3D object that suffer partial occlusions. We have evaluated the proposed descriptor for patch matching and keypoint matching, and found that it is more than competent when compared to the state-of-the-art descriptors. Further, we have conducted 3D reconstruction evaluations and found that the proposed method has produced significantly better results.


\begin{figure*}[t]
\begin{center}
\includegraphics[width=\linewidth]{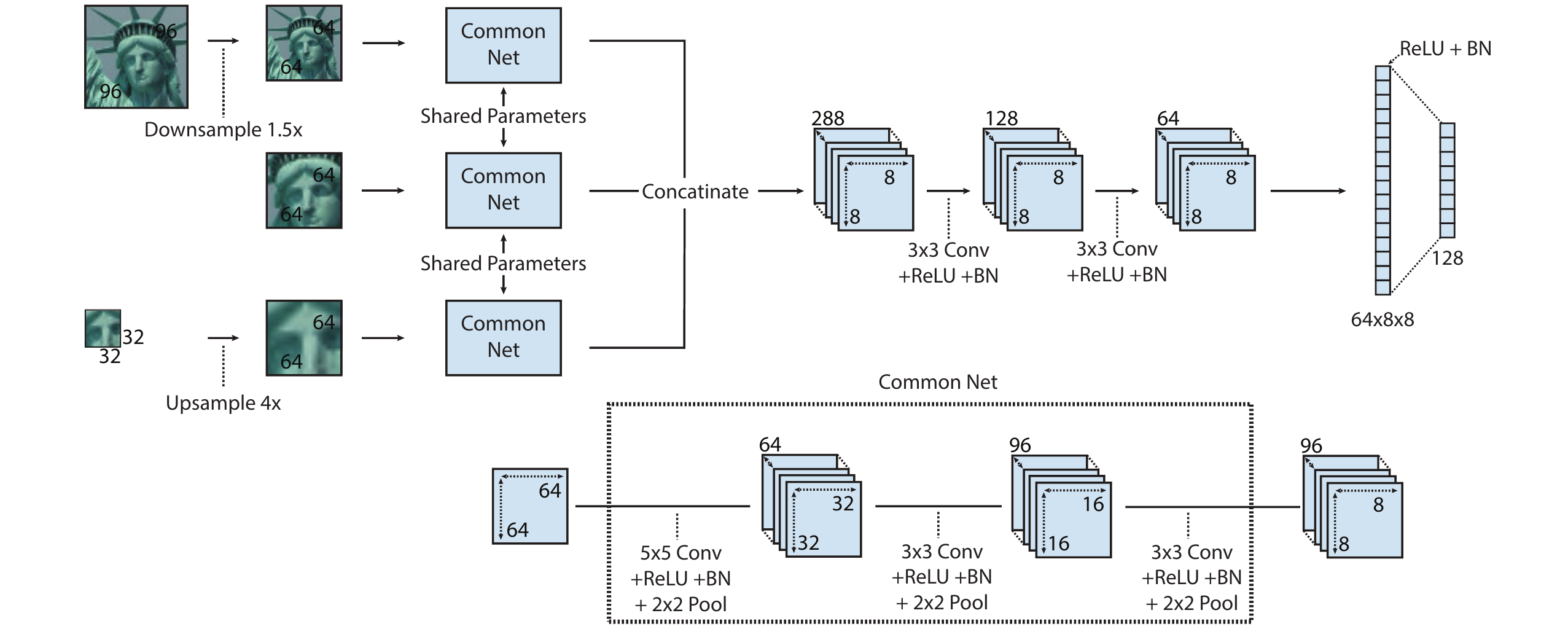}
\end{center}
\caption{\label{our-model} Illustration of the proposed network}
\end{figure*}

\subsection{Related Work}{\label{sec-rel-work}}
Several papers in the literature exist that address the challenges involved in designing image descriptors that are in turn used to find the image correspondences using local patch matching. These include the traditional hand-crafted descriptors such as SIFT~\cite{sift} and SURF~\cite{surf} and the more recent ConvNet based descriptors such as DeepDesc~\cite{deepdesc}, DeepCompare~\cite{deepcompare}, Matchnet~\cite{matchnet}, and Tfeat~\cite{tfeat}. Learning the descriptors for local patches using ConvNets was attempted early by Jahrer et al.~\cite{learned-desc} but was not followed up due to numerous practical issues and limited evaluation. However, with recent success of ConvNets and deep learning, Matching local image patches via learned descriptors became widespread study and many 
ConvNet based architectures have been proposed~\cite{deepcompare, matchnet, deepdesc,tfeat}. It has been shown in the literature that the descriptors learned using Siamese architecture based ConvNets considerably improve the matching performance ~\cite{deepcompare, matchnet, deepdesc}.

Few papers in the literature, study patch matching as a task~\cite{deepcompare,matchnet}, where the feature layers (Siamese network) and the metric learning layers (fully-connected layers) are jointly learnt in an end-to-end fashion. These type of ConvNets cannot be used as general descriptors for any tasks such as reconstruction except patch matching. Whereas,~\cite{deepdesc} uses the features extracted at the output of the Siamese networks without learning any non-linear decision network or metric learning layer. These type of descriptors, are generic in nature and can be used for many tasks as drop-in replacement of traditional descriptors including keypoint matching,  3D reconstruction, and tracking. Since, metrics to compare between patches are not learned, a generic metric such as $L_2$ distance to compare patches and train the network.  Learning feature descriptors using triplets of patches was investigated in~\cite{tfeat} using shallow networks in order to reduce the descriptor extraction time. Similar to~\cite{deepdesc,tfeat}, the aim of the proposed approach is to extract descriptors for local image patches that can be used for 3D reconstruction.

Inspired by the multi-bank architecture used in human-pose estimation~\cite{3bank-pose}, the proposed network uses a three bank network to encode scale variations of the image patches. Each bank shares common weights and hence the scaled patch inputs undergo similar transformation before being combined together and processed further. This helps the proposed network in being more robust to scale changes. Similar multi-resolution architecture has been proposed as a variant (central-surround two-stream model) in~\cite{deepcompare}. This multi-resolution model produces independent output combined by the metric learned layers. 
In the current literature, this type of architecture has not been studied for stand alone descriptors.

\section{Multi-Resolution Convolutional Neural Network}{\label{sec-model}} 

The Multi-Resolution Convolutional Neural Network which has the capability to capture better scale variance, we adapted it in a Siamese fashion ~\cite{siamese-lecunn} to learn patch descriptors of size $128$ dimensions. The proposed multi-bank network accepts image patches scaled to different resolutions, analogous to approximating the Laplacian pyramid for the input patch. The network has $3$ channels as shown in Fig.~\ref{our-model} and each channel accepts patches of size $64 \times 64$ pixels. The first channel takes  a patch of size $96 \times 96$ pixels downsampled to $64 \times 64$ pixels. The second channel takes a center-cropped $64 \times 64$ patch and the third channel takes $32 \times 32$ center-cropped patch scaled to $64 \times 64$. Each channel has identical structure consisting of three convolution layers and shares the parameters across the banks as shown in Fig.~\ref{our-model}. The output maps from the $3$ channels are then concatenated to form one bank and passed through $2$ convolution layers of $(128,\: 64)$ features respectively. The result is then flattened to form a $1$D tensor $(64 * 8 * 8 \:=\: 4,096)$ size and passed to a fully connected layer of $128$ dimensions.
\begin{figure*}[t]
\begin{center}
\includegraphics[width=\linewidth]{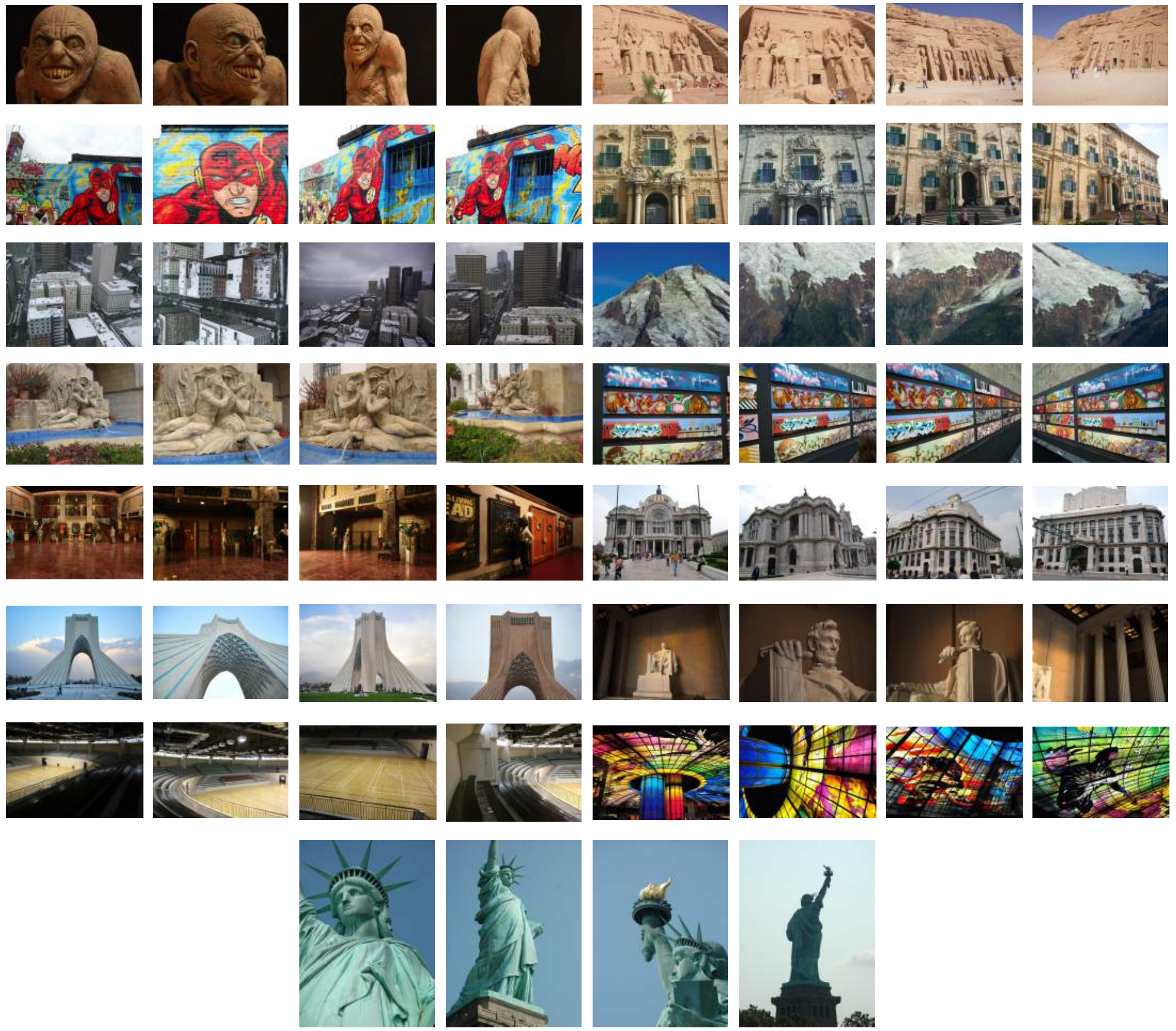}
\caption{\label{PS-multi-sample} Illustration of sample images from different scenes in the proposed PS dataset showing scenes with four view points each. As can be seen, these images have large variations in terms of baseline and pose.}
\end{center}
\end{figure*}
The output of this fully connected layer of trained network is used as the descriptors for the input image.

The Multi-Resolution architecture along with providing scale invariance also captures information at different extent. The top bank has a wider support region due to larger $96 \times 96$ patch, enabling the network to better distinguish among locally repeated patterns. In contrast, the bottom bank which is feed from up-sampling a a central $32 \times 32$ patch captures subtle changes which helps to discriminate from close by points. We also adopt a learned combination of the three outputs rather than leaving it to the point of mere concatenation as done in DeepCompare~\cite{deepcompare}.

Training of the network is performed in Siamese fashion using the contrastive loss function (Eq.~\ref{contrastive-loss}) as used in \cite{siamese-lecunn}. Here, $D_W$ is the output of the network whose parameters are $W$. $Y$ is a binary indicator function, whose value is $1$ when the pair $(A,B)$ forms a match and $0$ otherwise. The margin $m$, is the minimum distance by which a non-matching pair should be apart.
\begin{equation}{\label{contrastive-loss}}
\centering
L(W,\: Y,\: A,\: B) \: = \: \frac{1}{2}[Y(D_W)^2 + (1-Y)\lbrace \max(0, m - D_W) \rbrace^2]
\end{equation}

\section{The PS Dataset}{\label{sec-photo}}

In this paper we propose a new dataset, for learning generic descriptors, called the PhotoSynth-based dataset (PS). This dataset consists of two types of scenes, Multi-image and Single-image scenes.

{\bf Multi-image scene}: The scenes in this category focus on $3$D objects having distinctive edges. Each scene consists of $250$ color images on an average, and a corresponding sparse $3$D point cloud created using SFM~\cite{vsfm,Bundle-Adjustment}. Unlike the MVS dataset which has only $3$ scenes, the proposed dataset has $15$ scenes with considerable photometric and geometric variations. The number of patches per scene ranges from $75,000$ to $200,000$. Image patches are created by defining a square neighborhood around the projections of $3$D points in the images. The SFM process provides correspondences having wide baselines and large scale variations which cannot be obtained by stereo matching using handcrafted descriptors. Sample images of this category are illustrated in Fig.~\ref{PS-multi-sample}.

For a particular multi-image scene, let $P_{i}$ denote the set of all the patches belonging to a $3$D point $i$. The scale $s$, for a projection is given by the ratio $s \:=\: f / d$ and varies in range $(1,\: 2.0)$ in $P_{i}$. Here, $f$ is the focal length of the camera corresponding to the image the $3$D point is projected and $d$ is the distance between its camera center and the $3$D point projected along the camera's view direction. 
Viewpoint difference between a pair images is measured in terms of angle between their view directions. The viewpoint variation ranges from $10 - 60$ degrees. Square patches of size $96$ are cropped from images.

\begin{figure}[h]
\centering
\includegraphics[width=\linewidth]{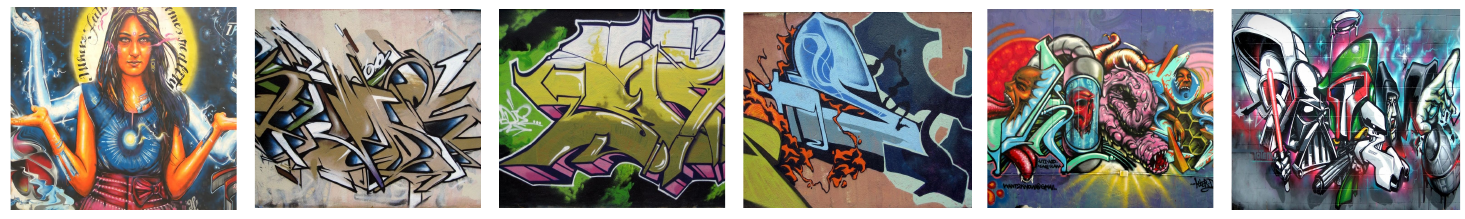}
\caption{\label{PS-single-sample} Reference images of different single-image scenes from the proposed PS dataset. }
\end{figure}

{\bf Single-image scene}: The scenes in this category contain images focusing a flat surface having varied textures, \eg, a wall. In such a scene, pairs are formed by taking a patch from the image and a random affine transform of the patch. Such transformations can be obtained dynamically while training the network. This process aids the training in two ways: (i) provide a wide variety of affine transforms between patches that are not present in the multi-image scenes (ii) It avoids over-fitting, since, the network sees the same patch taken from an image with different affine transform each time. Totally there are 10 images in this category. Sample images of this category are illustrated in Fig.~\ref{PS-single-sample}.

\begin{figure}
\centering
\includegraphics[width=0.48\linewidth]{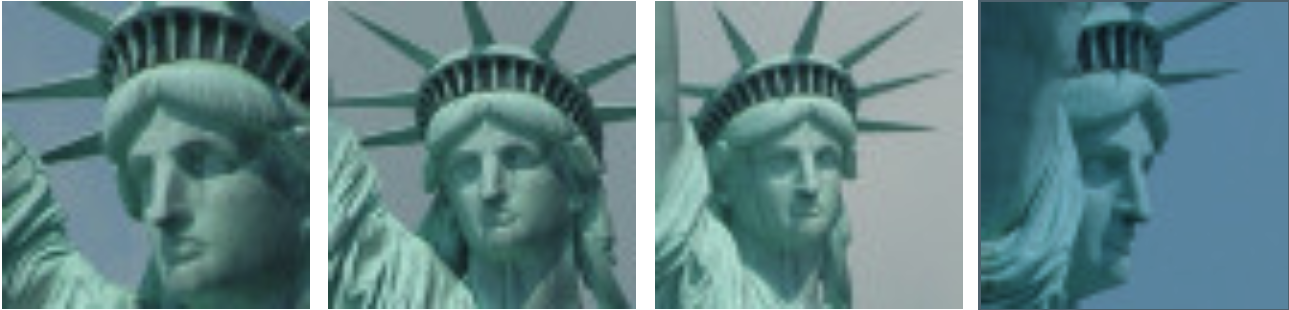} \hspace{0.01cm}
\includegraphics[width=0.48\linewidth]{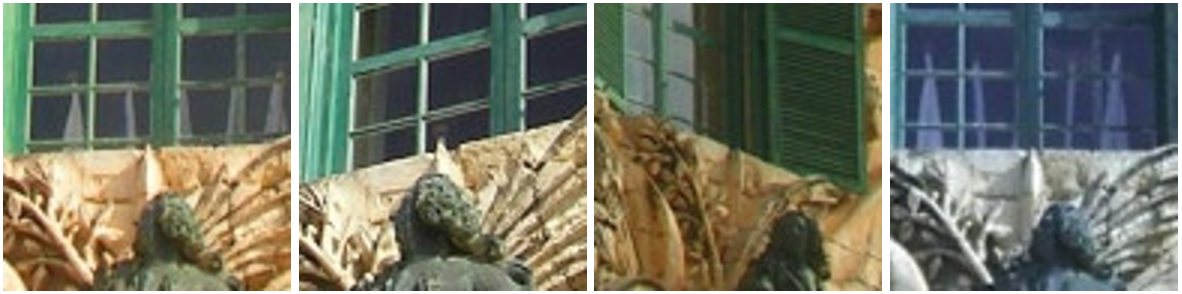} \\ \vspace{0.1cm}
\includegraphics[width=0.48\linewidth]{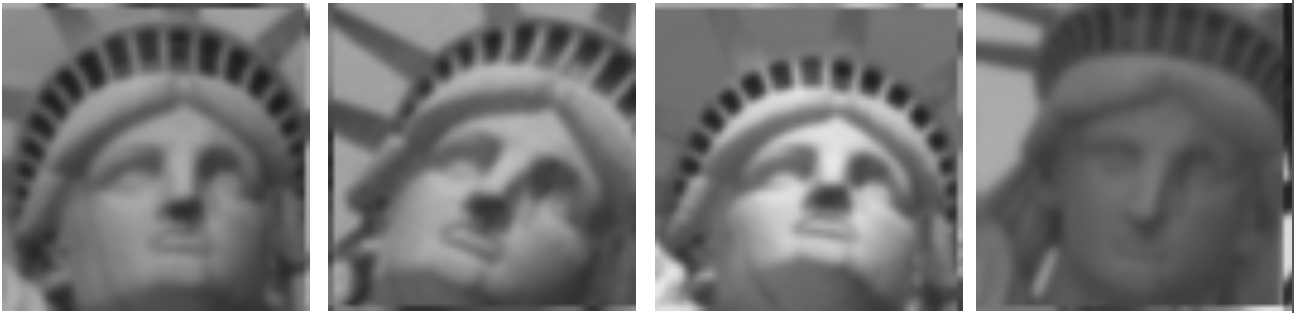} \hspace{0.01cm}
\includegraphics[width=0.48\linewidth]{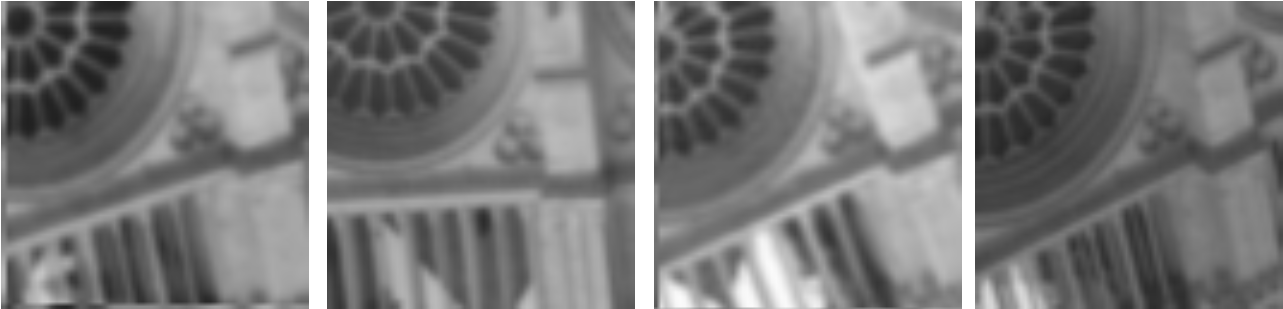}
\caption{\label{patchEx} Comparison between the proposed PS dataset (first row) and MVS dataset (second row). As can be seen, the patches depicting similar scenes in PS dataset have larger variation than that of MVS dataset.}
\end{figure}

The dataset in total contains 6 indoor scenes and 19 outdoor scenes. The resolution of the images are either $2000 \times 1300$ pixels or $1000 \times 750$ pixels. The format of our PS dataset\footnotetext{The dataset will be made publicly available} is similar to that of the MVS dataset. Each scene contains RGB patches, of size $96 \times 96$, Each scene is provided with a patch information list with an entry for all the patches in that scene and $3$D point index to which the patch is belongs to and the $(x,y)$ co-ordinates of the center of the patch in the grid image. Additionally, each scene contains a match-list containing pair of indices from the information list of all the matching pairs. The number of matching pairs in a scene varies from $60,000$ to $200,000$. For training we use $20$ scenes with $12$ and $8$ scenes from the \emph{multi-image} and \emph{single-image} categories respectively. The remaining $3$ \emph{multi-image} scenes and $2$ \emph{single-image} scenes form the test scenes. The test scenes have an additional list containing randomly selected $25,000$ matching and $25,000$ non-matching pairs. Sample patches from our dataset with comparison to MVS dataset are shown in Fig.~\ref{patchEx}.


\section{Experimental setup}{\label{exp-setup}}

The experimental setup for evaluating the proposed approach and for comparing with other approaches in literature is detailed in this section. The evaluation metrics and training methodology are described in sections~\ref{eval-metrics} and~\ref{net-training} respectively.

\subsection{Evaluation metrics}{\label{eval-metrics}}
Following \cite{deepdesc,tfeat}, matching score is used as metrics for patch matching. Matching score is the ratio of the number of correct predicted matches to the number of correspondences. Ground truth correspondences are computed using the homography associated with image pair. For a point in one image its nearest neighbor in the other image is predicted as a match.

We used \emph{\texttt{vl\_covdet}} from the \emph{\texttt{vl\_benchmark}} library~\cite{vl-feat} to extract patches and compute SIFT descriptors. The patches extracted by \emph{\texttt{vl\_covdet}} are affine normalized. 
We have also extracted unnormalized patches which provides a way to evaluate learnt descriptors without using scale and rotation information from the keypoint. 

To evaluate $3$D reconstructions via SFM using putative matching from different descriptors, \emph{DoG} keypoints are computed from \emph{\texttt{vl\_covdet}}. False positives are pruned by only selecting those pairs which form mutual nearest neighbors.
VisualSFM~\cite{vsfm} is used for reconstruction. Total number of points triangulated, average re-projection error and average track length (projections per $3$D point) are reported. 

\subsection{Training methodology}{\label{net-training}}
For training the proposed network, mini-batch gradient descent is used with batch size of $64$ pairs and $1000$ batches per epoch. Each batch contains $16$ matching and $48$ non-matching pairs. Further, the matching pairs in a batch are systematically distributed in $4$ ranges such as $\left[0, 0.4 m \right], \left[0.4, 0.8 m \right], \left[0.8, 1.2 m \right], \left[1.2, 1.6 m \right] $ in a ratio of $4:4:6:2$. Here, $m$ is the margin of contrastive loss (set to $2$). For proper training we have used negative mining strategy, where few wrongly classified pairs in an epoch is used for training the subsequent epochs.

Similar to matching pairs, all non-matching pairs were also divided into four ranges based on margin distance.  A subset $S$ of patches is taken from all the $N$ patches. For every patch $p$ in $S$, we divide $N$ patches into $4$ buckets (first one containing the closest patches and last one containing the farthest patches from $p$) and sample $6$ patches from the first two buckets in the ratio $4:2$ and form non-matching pairs with $p$. Care is taken to ensure that none of the matching patches are paired as non-matching. We don't look beyond the first $2$ buckets as it has been observed that after the first epoch the distance of non-matching pairs lying in the $3$rd and $4$th bucket are above the margin $m$ and don't contribute to the gradients. 

To reduce over-fitting and achieve rotation and scaling invariance, the patches are perturbed randomly during training. The perturbations include rotating and scaling the patch with random values within the range $\left[ -\pi/8,\: +\pi/8 \right]$ and $\left[ 1.0,\: 1.1 \right]$ respectively. Perturbations are also used to create matching and non-matching pairs from the single image scenes. A matching pair is formed by pairing a patch and an affine transformation of it. For non-matching pairs, a patch is paired with an affine transformation of some other patch from the same single image. 

\section{Results}{\label{sec-results}}
In this section, we evaluate the performance of the proposed approach on patch pair classification, keypoint matching and $3$D reconstruction tasks and compare with the recent approaches in literature. The patch pair classification task is to classify a given pair of patches as matching or non-matching. Though in the real-world this type of classification is not feasible, we report the performance for completeness. The keypoint matching task is to find matching patches around keypoints detected in images captured from different views. The results of pair classification, keypoint matching and 3D reconstruction are reported below.

\subsection{Patch pair classification}{\label{patch-pair-res}}
The MVS dataset~\cite{mvs-new} is used to measure the ability of a descriptor to discriminate positive pairs of patches from negative pairs. It has $3$ scenes, Liberty (Lib), Notredame (Not) and Yosemite (Yos) with $450,092$, $468,159$ and $633,587$ patches respectively. Each scene is also provided with a list of pairs with $50\%$ matching and $50\%$ non-matching pairs.
Approaches in ~\cite{matchnet,deepcompare,tfeat} use the evaluation mentioned in~\cite{std-eval} where model training is based on single scene. However, in~\cite{deepdesc}, training is based on two scenes and tested on remaining one. The evaluation is performed by thresholding distance scores between patch pairs on ROC curve. The results are shown in table ~\ref{table:MVS-1-comparison}. The numbers reported in the table is the false positive rate at $95\%$ true positive rate (FPR95). It is observed that the model trained on single scenes performs marginally lower than~\cite{tfeat} only in some cases. Since our model capacity is intentionally made large (in order to achieve descriptor generalization using the proposed PS dataset) and with the MVS dataset being small, the problem of over-fitting is observed when training on single scenes. However, it should be noted that the pair classification is not of practical importance when compared to the keypoint matching and it is only reported for completeness.

\begin{table*}[ht]
\centering
\begin{tabular}{l c c c c c c c c c c c}
\hline
Training & & Not \hspace{-0.7cm} & & \hspace{-0.7cm} Lib & Not \hspace{-0.7cm} & & \hspace{-0.7cm} Yos & Yos \hspace{-0.7cm} & & \hspace{-0.7cm} Lib \\
Testing & & & Yos & & & Lib & & & Not \\
\hline
Descriptor & \# & & & & & & & & & & mean \\
\hline
Sift~\cite{sift} & 128 & & 27.29 & & & 29.84 & & & 22.53 & & 26.55\\
DeepCompare {\tiny \emph{siam}}~\cite{deepcompare} & 256 & 15.89 \hspace{-0.7cm} & & \hspace{-0.7cm} 19.91 & 13.24 \hspace{-0.7cm} & & \hspace{-0.7cm} 17.25 & 8.38\hspace{-0.7cm} & & \hspace{-0.7cm} 6.01 & 13.45\\
DeepCompare {\tiny \emph{siam2stream}} & 512 & 13.02 \hspace{-0.7cm} & & \hspace{-0.7cm} 13.24 & 8.79 \hspace{-0.7cm} & & \hspace{-0.7cm} 12.84 & 5.58\hspace{-0.7cm} & & \hspace{-0.7cm} 4.54 & 9.67\\
DeepDesc~\cite{deepdesc} & 128 & & 16.19 & & & 8.82 & & & 4.54 & & 9.85\\
Matchnet~\cite{matchnet} & 512 & 11 \hspace{-0.7cm} & & \hspace{-0.7cm} 13.58 & 8.84 \hspace{-0.7cm} & & \hspace{-0.7cm} 13.02 & 7.7\hspace{-0.7cm} & & \hspace{-0.7cm} 4.75 & 9.82\\
TFeat {\tiny \emph{margin${}^{*}$}}~\cite{tfeat} & 128 & {\bf 7.08} \hspace{-0.7cm} & & \hspace{-0.7cm} {\bf 7.82} & {\bf 7.22} \hspace{-0.7cm} & & \hspace{-0.7cm} {\bf 9.79} & {\bf 3.85}\hspace{-0.7cm} & & \hspace{-0.7cm} {\bf 3.12} & {\bf 6.47}\\
\emph{Proposed}& 128 & 13.8\hspace{-0.7cm} & & \hspace{-0.7cm}13.9 & 7.4\hspace{-0.7cm} & & \hspace{-0.7cm}11.4 & 5.01\hspace{-0.7cm} & & \hspace{-0.7cm} 3.49 & 9.16\\
\hline
\end{tabular}
\vspace{1em}
\caption{\label{table:MVS-1-comparison} Comparison of the state-of-the-art results on the MVS dataset with the proposed model in terms of $FPR95$ (lower is better). Yos:Yosemite, Lib:Liberty, Not:Notredame}
\end{table*}

\vspace{-0.1cm}

\subsection{Keypoint matching}
\begin{figure}[h]
\centering
\includegraphics[width=0.9\linewidth,height=0.4\linewidth]{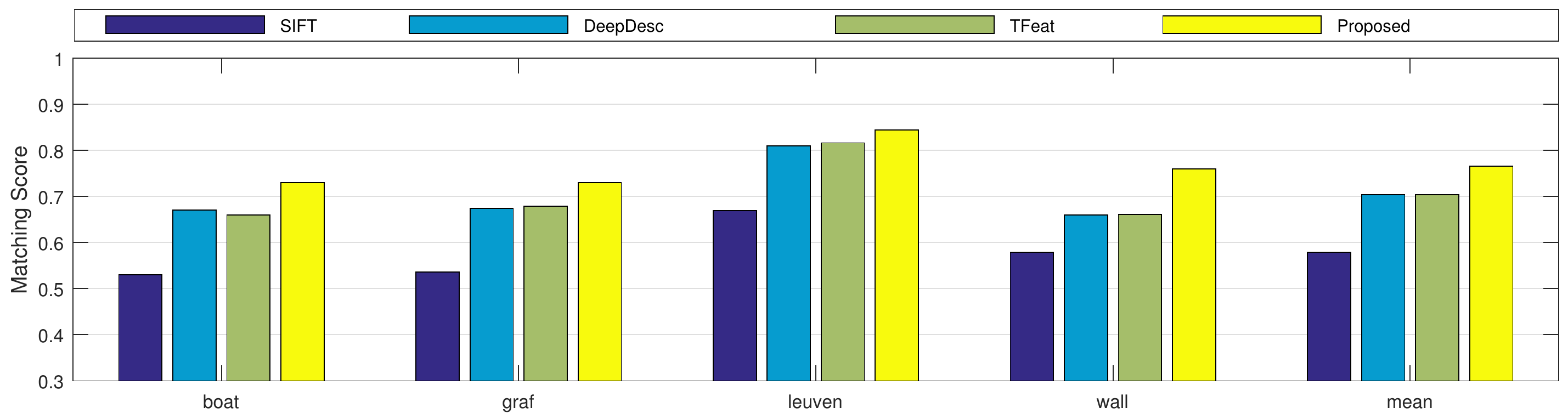}
\caption{\label{fig-oxford-norm} Comparison of proposed descriptor against SIFT, DeepDesc and TFeat on $4$ scenes in Oxford~ACRD dataset using normalized patches obtained from Harris-Affine keypoints using MScore metric}
\end{figure}

\begin{figure}[h]
\centering
\includegraphics[width=\linewidth,height=0.4\linewidth]{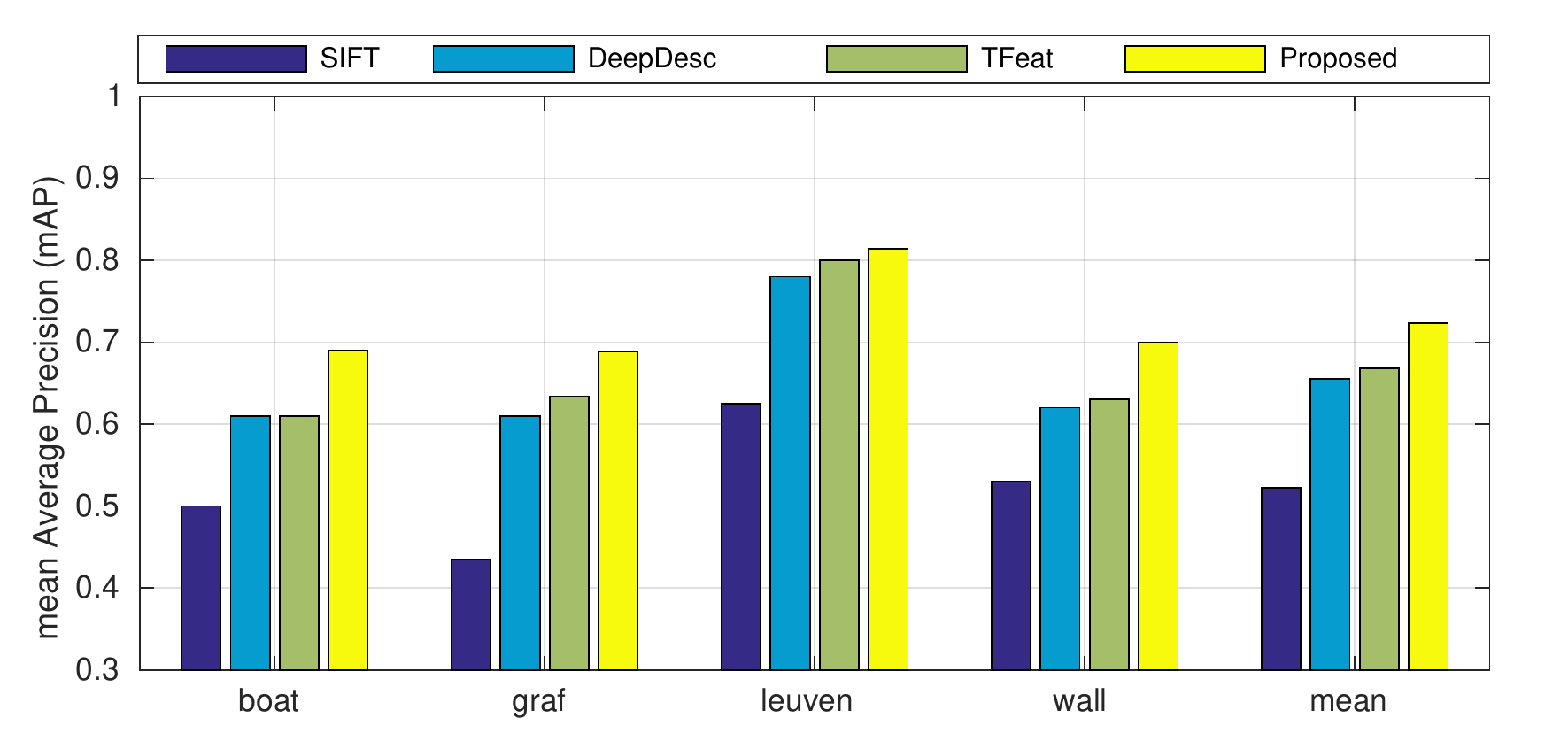}
\caption{\label{fig-oxford-map}  Comparison of proposed descriptor against SIFT, DeepDesc and TFeat on $4$ scenes in Oxford~ACRD dataset using normalized patches obtained from Harris-Affine keypoints using mAP metric.}
\end{figure}

{\label{keypoint-res}}
The Oxford~ACRD~\cite{oxford} and SG dataset~\cite{generated-dataset} are used for evaluating the keypoint matching performance of different descriptors. The Oxford~ACRD contains real images with different geometric and photometric transformations for different scene types. We consider four scenes: boat (zoom, rotation), graffiti (viewpoint), leuven (light) and wall (viewpoint). 
As mentioned in section~\ref{exp-setup} we use matching score (MScore) and mAP values as metrics. Fig.~\ref{fig-oxford-norm} and Fig.~\ref{fig-oxford-map} shows the MScore comparison on $4$ scenes from the Oxford~ACRD for normalized patches obtained using Harris-Affine keypoints. As can be observed, the proposed descriptor outperforms all the other descriptors on all the scenes.

\begin{figure}[h]
\centering
\includegraphics[width=0.98\linewidth, height=0.5\linewidth]{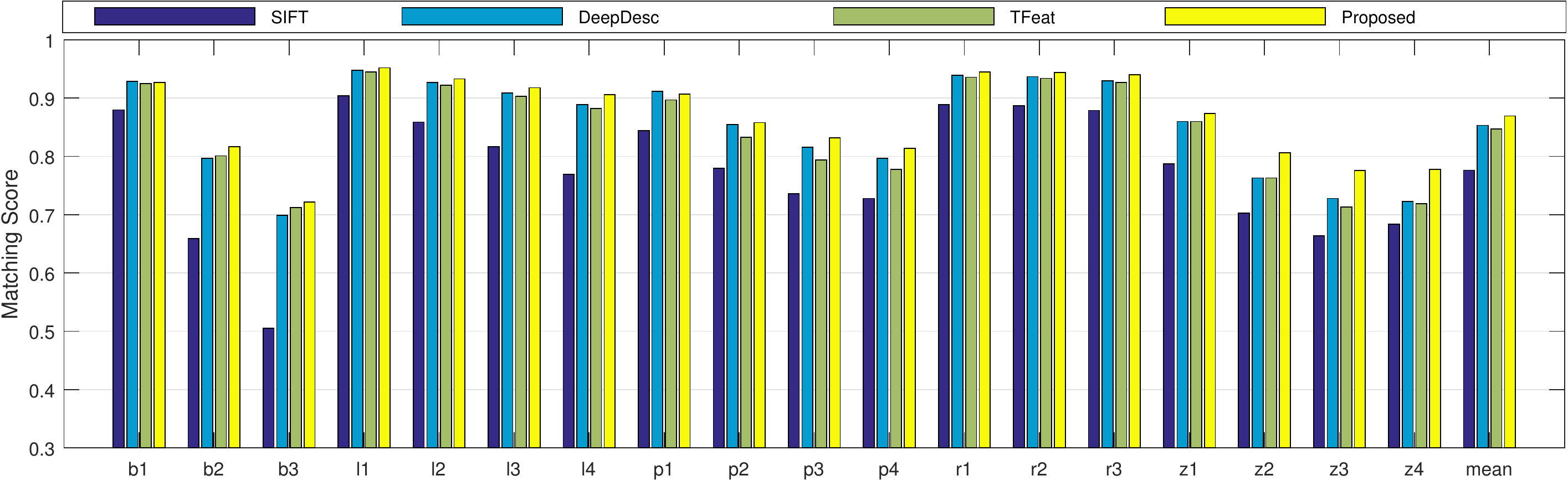}
\caption{\label{fig-synth-norm} Comparison of proposed descriptor against SIFT, DeepDesc and TFeat on Synthetic dataset for normalized patches using Harris-Affine keypoints. Comparison for $5$ different transformations ($b$-blur, $l$-lighting, $p$-perspective, $r$-rotation and $z$-zoom) are shown. Numbers next to the transformations indicate the degree of transformation.}
\end{figure}

\begin{figure}[h]
\centering
\includegraphics[width=0.98\linewidth, height=0.5\linewidth]{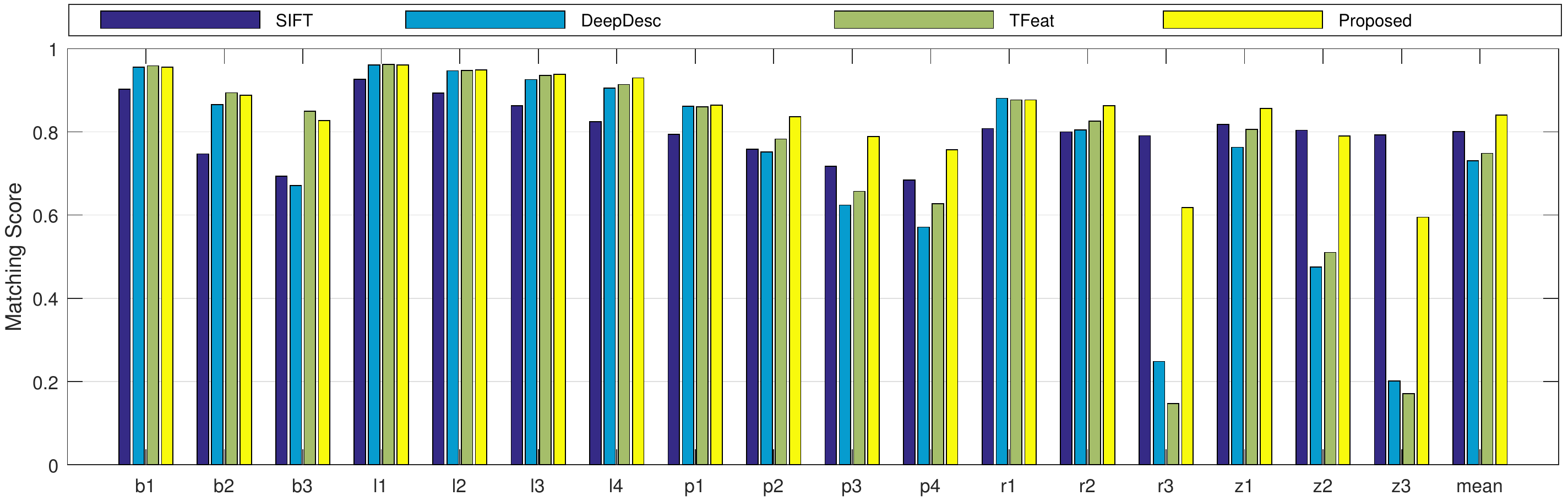}
\caption{\label{fig-synth-unnorm} Comparison of proposed descriptor against SIFT, DeepDesc and TFeat on Synthetic dataset for unnormalized patches. Notations similar to those in fig~\ref{fig-synth-norm}}
\vspace{-0.2cm}
\end{figure}

\begin{figure}[h]
\centering
\includegraphics[width=\linewidth, height=0.5\linewidth]{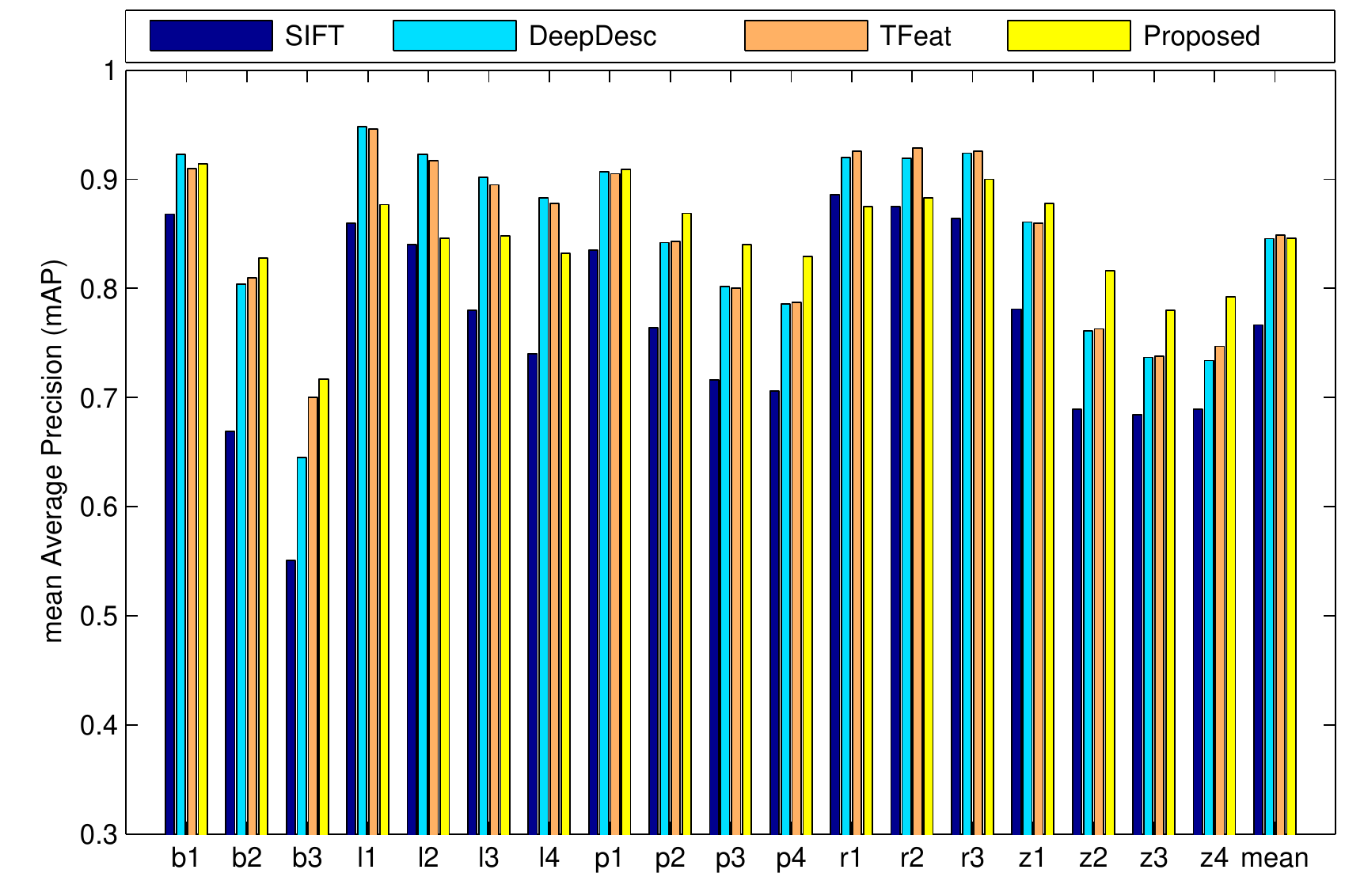}
\caption{\label{fig-synth-map} Comparison of proposed descriptor against SIFT, DeepDesc and TFeat on Synthetic dataset for normalized patches using Harris-Affine keypoints using mean Average Precision (mAP). Comparison for $5$ different transformations ($b$-blur, $l$-lighting, $p$-perspective, $r$-rotation and $z$-zoom) are shown. Numbers next to the transformations indicate the degree of transformation.}
\end{figure}

The SG dataset has $16$ scenes with each scene represented by a reference image. For each scene, the reference image is synthetically warped geometrically and photometrically to generate new images. The transformations include \emph{blur}, \emph{lighting}, \emph{rotation}, \emph{zoom} (scaling), \emph{perspective} (viewpoint). Fig.~\ref{fig-synth-norm} and Fig.~\ref{fig-synth-unnorm} show MScore of different descriptors on the SG dataset~\cite{generated-dataset} for normalized patches (using Harris-Affine keypoints) and unnormalized patches respectively. The plots show comparison for different degrees of $5$ transformations in the dataset. For each transformation, as the degree of variation increases, the performance of the proposed descriptor is observed to be better than the other descriptors being compared against. For the unnormalized patches, even though SIFT is better for large values of zoom and rotation, proposed descriptor is better for all other transformations and is always better or comparable to DeepDesc and TFeat. Thus, the proposed descriptor is robust to different transformations and can handle large variations better than all the other descriptors. In Fig.~\ref{fig-synth-map}, comparison of mAP for the SG dataset is shown.

We observe that our descriptor outperforms all the other descriptor in the Oxford-ACRD for all scenes and majority of the transformations in the SG dataset. We observe that for low geometric transformation in the SG dataset \eg 'lighting', we have inferior mAP values compared to DeepDesc and Tfeat although having similar Matching Score (as shown in Fig.~6 in the main paper). One possible reason is that our PS dataset which is used for training have much more difficult matching and non-matching pairs than MVS. This makes distance between matching and non-matching pair is much more spread out and have lower mAP which is based on threshold over descriptor distances.

\vspace{-0.2cm}

\subsection{3D reconstruction}{\label{3d-res}}
In this section, we compare reconstructions using putative matches obtained using our model, DeepDesc, Tfeat and SIFT. We use the \emph{fountain-P11},  \emph{herz-Jesu-P8} and entry-P10 datasets from~\cite{Strecha} to reconstruct 3D points using SFM. 
The metrics used for evaluation are discussed in Sec.~\ref{eval-metrics}.
Table~\ref{recon-eval} shows the results of reconstructions obtained using different descriptors. 

\begin{table*}[t]
\centering
\begin{tabular}{l c c c c c}
\hline
& Dataset & SIFT & DeepDesc & Tfeat & Proposed\\
\hline
No. Points & F-P11 & 25.6k & 23.5k & 25.4k & {\bf 26.0k}\\
& HJ-P8 & 11.0k & 11.0k & 11.4k & {\bf 11.5k}\\
& E-P10 & 14.8k & 13.8K & 14.2k & {\bf 16.0k}\\
\hline
Reproj. Err & F-P11 & {\bf 1.93} & 3.40 & 2.99 & 2.64\\
& HJ-P8 & {\bf 3.76} & 6.18 & 5.17 & 4.46\\
& E-P10 & {\bf 4.02} & 8.02 & 4.59  & 4.29\\
\hline
Tot. Proj. & F-P11 & 98.3k & 87.8k & 97.2k & {\bf 101.3k}\\
& HJ-P8 & 36.9k & 35.8k & 38.7k & {\bf 40.3k}\\
& E-P10 & 57.6k & 50.9k & 55.5k & {\bf 64.9k}\\
\hline
Avg. track len. & F-P11 & 3.83 & 3.73 & 3.82 & {\bf 3.90}\\
& HJ-P8 & 3.36 & 3.36 & 3.39 & {\bf 3.50}\\
& E-P10 & 3.88 & 3.67  & 3.90 & {\bf 4.04} \\
\hline
Inlier Matches. & F-P11 & 176K & 164K & 188K & {\bf 197K}\\
& HJ-P8 & 51K & 53K & 60K & {\bf 63K}\\
& E-P10 & 111K & 106K  & 119K & {\bf 145K} \\
\hline
\end{tabular}
\vspace{1em}
\caption{\label{recon-eval} Reconstruction results using different descriptors on 3 datasets from \emph{Strecha}. $4$ metrics (detailed in section~\ref{eval-metrics}) are used. Higher is better for Number of 3d points, Total projections and Average track length metrics while lower is better for re-projection error}
\end{table*}

\begin{figure}[h]
\centering
\includegraphics[width=\linewidth,height=0.9\linewidth]{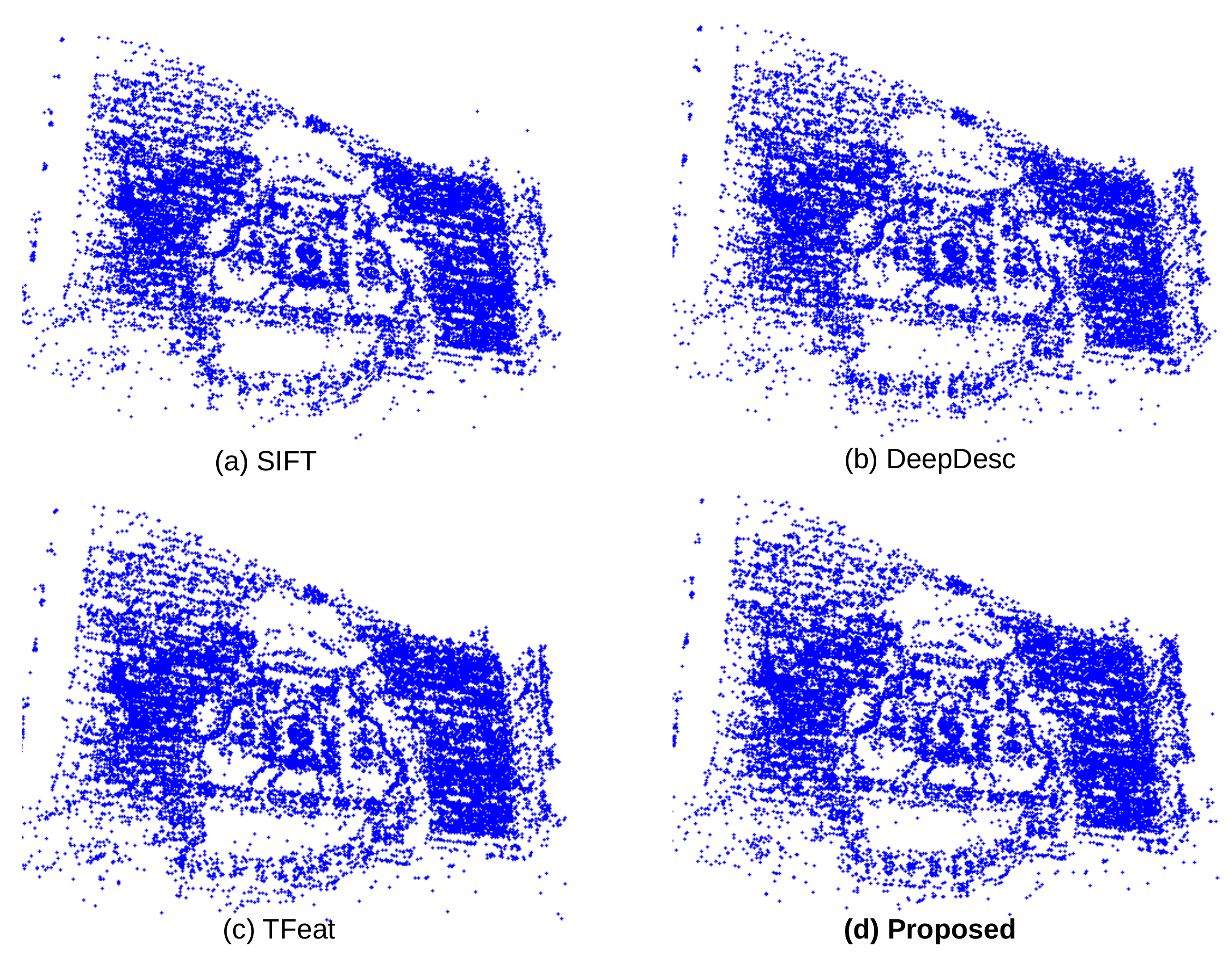}
\caption{\label{fount-3d} 3D reconstruction comparison of proposed descriptor against SIFT, DeepDesc and TFeat on \emph{fountain-P11} scene. The bottom part of the fountain is better reconstructed for the proposed descriptor.}
\end{figure}

\begin{figure}[h]
\centering
\includegraphics[width=1\linewidth,height=0.9\linewidth]{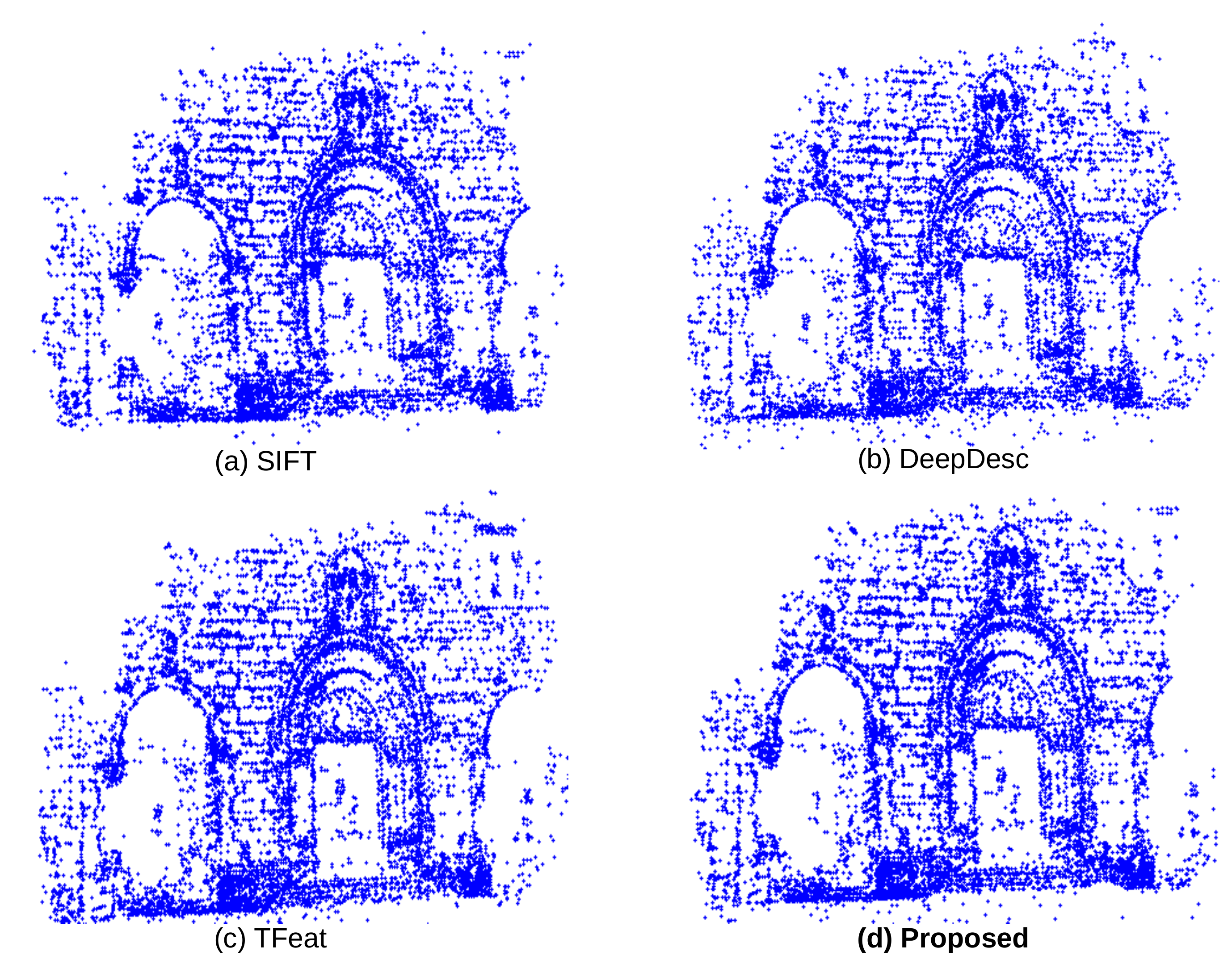}
\caption{\label{herz-3d} 3D reconstruction comparison of proposed descriptor against SIFT, DeepDesc and TFeat on herz-Jesu-P8 scene.}
\end{figure}

From table~\ref{recon-eval}, we observe that our proposed model performs better  than DeepDesc~\cite{deepdesc} and TFeat~\cite{tfeat} on all the four metrics considered. In comparison to SIFT, the proposed descriptor is better on three of the four metrics. Even though the re-projection error is higher for the proposed descriptor when compared to SIFT, the number of inlier matches between different views is higher for our descriptor along with the average track length (measure of number of projections of a $3$D point in different views). $3$D reconstructions for the \emph{fountain-P11} and \emph{herz-Jesu-P8} scenes of \emph{Strecha} are shown in Fig.~\ref{fount-3d} and Fig.~\ref{herz-3d} respectively. We observe that all methods produce visually indistinguishable results in most parts of the reconstructions. However in Fig.~\ref{fount-3d}, the bottom part of the fountain reconstruction is better for the proposed descriptor.

Thus, the proposed descriptor is better than all the other descriptors for reconstruction task especially among the learned descriptors.

\vspace{-0.2cm}
\subsection{Efficiency}{\label{result-eff}}
We have used Nvidia-Titan X for training and testing. On a batch size of 128 averaged over 1000 batches, the forward propagation times of Tfeat, DeepDesc and our network are 3.5, 175 and 14 micro-seconds respectively. We observe that our network is slightly slower than Tfeat though having multi-resolution banks and operating on larger patch sizes.  
\vspace{-0.2cm}

\section{Conclusion}
In this paper, we proposed a learning based local image descriptors for patch matching and 3D reconstruction. For designing efficient learning based descriptors using ConvNet a good combination of dataset as well as architecture is important. We propose the use of multi-resolution architecture and we have introduced a new dataset with $25$ scenes of varied content and containing images with high geometric transformations. With training, ConvNet with our dataset  to obtain descriptors we have found that it is invariant to geometric changes than other learned descriptors when key-point information is not used. We have also found that it generated on average $5$\% more number of points when compared to other descriptors during reconstructions. The proposed combination has also produced increased image coverage per point on wide baseline scenes. With these results, we can conclude that the proposed combination of multi-resolution ConvNet with the new dataset produces descriptor that generalizes across the dataset.

{\small
\bibliographystyle{ieee}
\bibliography{egbib}
}

\end{document}